\pdfoutput=1

\documentclass[11pt]{article}

\usepackage[preprint]{acl}

\usepackage{times}
\usepackage{latexsym}

\usepackage[T1]{fontenc}

\usepackage[utf8]{inputenc}

\usepackage{microtype}

\usepackage{inconsolata}

\usepackage{graphicx}

\usepackage{enumitem}

\title{PerQ: Efficient Evaluation of Multilingual Text Personalization Quality}

\author{Dominik Macko \and Andrew Pulver \\
  Kempelen Institute of Intelligent Technologies \\
  \texttt{dominik.macko@kinit.sk}, \texttt{andrew.pulver@intern.kinit.sk} \\}

\begin{document}
\maketitle
\begin{abstract}
Since no metrics are available to evaluate specific aspects of a text, such as its personalization quality, the researchers often rely solely on large language models to meta-evaluate such texts. Due to internal biases of individual language models, it is recommended to use multiple of them for combined evaluation, which directly increases costs of such meta-evaluation. In this paper, a computationally efficient method for evaluation of personalization quality of a given text (generated by a language model) is introduced, called PerQ. A case study of comparison of generation capabilities of large and small language models shows the usability of the proposed metric in research, effectively reducing the waste of resources.
\end{abstract}

\section{Introduction}
\label{sec:intro}

Due to lack of existing metrics for various aspects of qualitative analysis of the texts and the time-intensive human evaluation (especially challenging in multilingual settings), the researchers often result in using general-purpose large language models (LLMs) for annotation purpose. Given the scope and variety of applied research tasks, such usage of highly resource-intensive LLMs represents huge waste of resources around the globe. Therefore, resource-efficient methods must be developed to evaluate specific aspects of the texts.

This work is focused on evaluation of text personalization quality. The personalization refers tailoring the text for specific target (e.g., user group, distribution platform). Since there is no standard metric to evaluate this aspect, we propose a new metric (PerQ) that is able to reflect the majority judgment of three diverse LLMs (various sizes and architectures) in efficient manner\footnote{\url{https://anonymous.4open.science/r/PerQ}}.

The contributions of this work are as follows:
\begin{itemize}[noitemsep,topsep=0pt]
    \item the \textbf{proposal of a new metric} for resource-efficient estimation of the text personalization quality,
    \item the \textbf{proposal of a methodology for metric training} that is easily extensible to train new metrics for various aspects of a text,
    \item the \textbf{comparison and evaluation of multiple base models} that offer different tradeoffs between performance and inference costs,
    \item the \textbf{execution of a case study} to showcase usefulness of the proposed metric in research.
\end{itemize}

\begin{figure*}[!t]
\centering
\includegraphics[width=\linewidth]{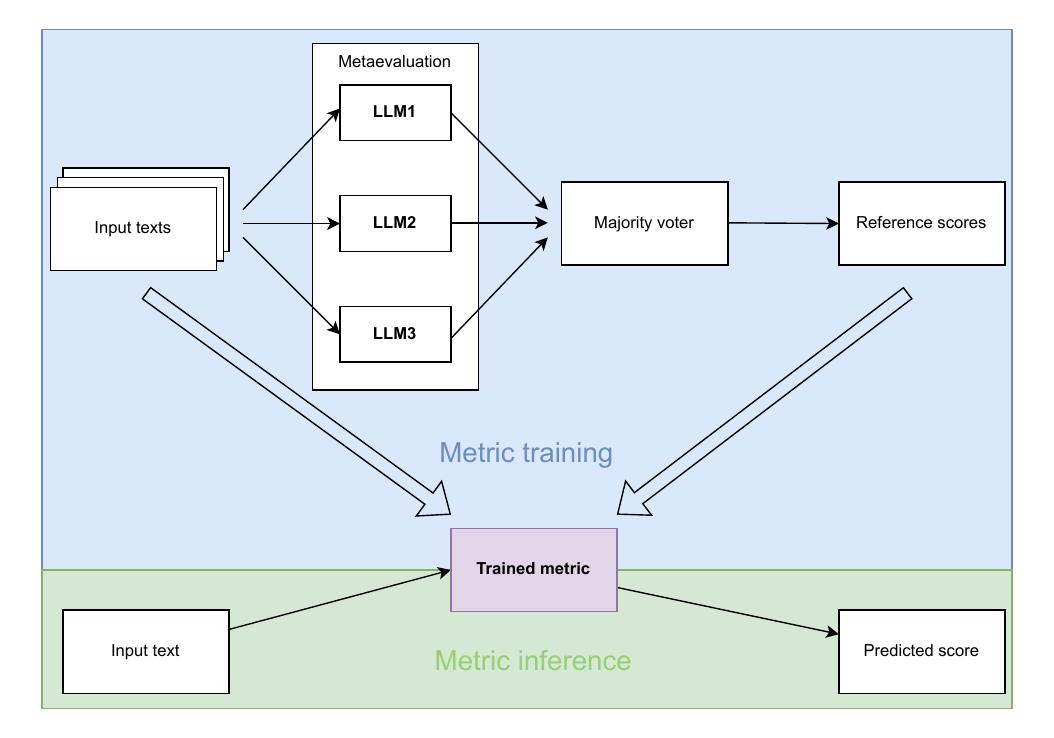}
\caption{Framework of the proposed metric training and inference.}
\label{fig:framework}
\end{figure*}

\section{Related Work}
\label{sec:related}

Traditional metrics, such as BLEU, ROUGE, or BERTScore require reference texts and measure similarity towards those; thus, limiting their utilization to evaluate purely generated content. Such metrics have been used to evaluate personalization (based on user history profile), for example, in LaMP \citep{salemi-etal-2024-lamp}, identifying the need for new metrics. The AuPEL method \citep{wang2023automatedevaluationpersonalizedtext} uses LLMs as the evaluators of personalized text generation and shows that is superior to the traditional text similarity metrics. The ExPerT framework \citep{salemi2025experteffectiveexplainableevaluation} also utilizes LLMs to extract atomic aspects from the generated and reference texts and compare them based on content and writing style.
The PREF framework \citep{fu2025prefreferencefreeevaluationpersonalised} does not rely on reference texts to evaluate personalized text quality. The personalization is rather focused on a user preference, i.e. user-specific alignment. In PerDisNews study \citep{zugecova-etal-2025-evaluation}, the personalization quality has been evaluated using three LLMs in zero-shot manner (without reference texts) in order to limit LLMs internal biases. Such approach has shown strong correlation with human judgment.

The LLMs are nowadays commonly used for evaluation of various aspects of a text, such as linguistic quality, coherence, hallucinations, even in multilingual manner \citep{chang2025exploringmultilingualnlgevaluation, hada-etal-2024-metal, macko-etal-2025-multisocial}.
However, the reliance on huge LLMs (even multiple inferences for a single text) to evaluate personalization quality (or other aspects of a text) is infeasible in many cases, especially for exploratory research, where quick feedback is required to continue. Therefore, new easily usable and efficient metrics to evaluate quality of personalization are needed.

\section{The Proposed Evaluation Metric}
\label{sec:method}

To cope with the existing challenges, we propose a reference-free metric to evaluate personalization quality based on smaller fine-tuned language model. It is trained as a classifier based on majority-voted scores of three different LLMs. The diverse combination of LLMs reduces internal biases of individual LLMs (as suggested by \citealp{zugecova-etal-2025-evaluation}). It estimates a score similarly as the reference-free version of the COMET metric \citep{rei-etal-2021-references}, used nowadays for evaluation of machine translation.

The LLMs are used only once to score a sufficient number (recommended at least 100 samples per class) of texts. Afterwards, the metric is trained and provides the score predictions with sufficient accuracy at considerably lower costs (e.g., result obtained in seconds in comparison to hours). It can be directly reused for the same scoring scale on different data without further training.

As illustrated in Fig.~\ref{fig:framework}, in the inference stage (i.e., metric usage), only the input text is provided, and the trained metric predicts the score regarding the personalization quality of the given text. The quality score reflects the trained classes based on scoring scale (e.g., class 0, class 1, class 2). The proposed approach for metric development can be reused for various scoring scales as well as for various text analysis metrics (e.g., general text quality, fluency, coherency). This is useful especially for multilingual evaluation, where usually non-English text metrics are heavily not available.

\begin{table*}[!t]
\centering
\small
\resizebox{\linewidth}{!}{
\begin{tabular}{p{3cm}p{12cm}}
\hline
\textbf{Score} & \textbf{Description} \\
\hline
0 - Not personalized at all to the platform &
The text shows no adaptation to the stylistic or structural norms of the platform.
a) The format is incompatible with the platform (e.g., long-form text posted to a platform that favors brevity like X/Twitter).
b) The tone is formal or generic and does not reflect the platform's communication style.
c) There is no use of features native to the platform (e.g., hashtags, tags, emojis, memes, short videos, interactive polls).
d) The content could be copied and pasted onto any platform with no change in impact.
\\
\hline
1 - Low personalization to the platform &
The text fits on the platform but lacks adaptation in tone, format, or engagement style.
a) The post length fits within the platform limits but does not leverage platform-specific formatting (e.g., uses no line breaks or bullets on LinkedIn).
b) The tone is only slightly adjusted (e.g., slightly more casual for Instagram but still generic).
c) Minimal or generic use of platform features (e.g., hashtags that are too broad or not trend-specific).
d) Content could be slightly edited for another platform with minimal effort.
\\
\hline
2 - Moderately personalized to the platform &
The text shows clear effort to fit the platform's tone and structure, though some elements could be more naturally integrated.
a) Includes relevant platform features (e.g., a few well-chosen hashtags, a meme format, a call to action) but doesn't use them to their full potential.
b) The tone is mostly aligned with the platform but may feel slightly off or inconsistent in parts.
c) The format is generally appropriate for the platform, but could be better optimized (e.g., a caption on Instagram that is a bit too long, or a thread on X that doesn't build momentum).
d) Engagement prompts are present but not especially compelling or natural for the platform's audience.
\\
\hline
3 - Well-personalized to the platform &
The text is naturally and effectively tailored to the specific platform's norms, tone, and engagement style.
a) Uses platform-specific features effectively (e.g., carousel for LinkedIn or Instagram, threads on X, storytelling formats on TikTok).
b) Tone is fully aligned with what the platform's audience expects (e.g., professional yet authentic on LinkedIn, snappy and meme-savvy on X, visual and emotive on Instagram).
c) Leverages platform-specific engagement cues (e.g., “comment below,” “share with a friend,” “stitch this”).
d) The message is concise, formatted appropriately, and makes full use of the platform's strengths.
e) Likely to generate high engagement because of its alignment with the platform's user behavior and content culture.\\
\hline
\end{tabular}
}
\caption{Scoring schema for evaluation of personalization quality.}
\label{tab:scoring_schema}
\end{table*}

\subsection{PerQ}
\label{sec:perq}

The research question targeted by PerQ metric development is (\textbf{RQ1}) \textit{``Can an evaluation metric been trained to score a text personalization quality with the accuracy above chance?''}.

For training of the PerQ metric, a dataset of multilingual texts personalized for various targets is required, containing sufficient amount of various personalization quality scores. Since no such dataset is available, we have created a new dataset by using multilingual news articles from MassiveSumm \citep{varab-schluter-2021-massivesumm}.

We have pseudo-randomly selected 100 samples per each of the 7 selected languages (English, German, French, Italian, Slovak, Russian, and Hungarian). The titles (headlines) of the articles have been used for personalized text generation and the contents of the articles have been used for personalization of the existing text (modification, adjustment). As targets, we have selected 3 social-media platforms (Twitter/X, Telegram, and Signal). Finally, we have used smaller and larger variants of 3 LLM architectures (Gemma-3, Qwen3, Llama-3) to generate the personalized texts. In this way, we have ended up with sufficient amount and variety of different texts of various quality. Using all combinations of the above mentioned 7 languages, 2 personalization types, 3 platforms, and 6 LLMs, resulted in 25,200 generated (possibly personalized) texts.

After the texts have been generated, we have used 3 different LLMs (Mistral-Small-3.1-24B-Instruct-2503, Aya-Expanse-32B, and QwQ-32B) to metaevaluate the quality of the texts by providing a detailed scoring schema (see Table~\ref{tab:scoring_schema}). Using majority voting, the received scores have been modified to a single score per text (the lowest score used when majority was indecisive -- i.e, each LLM predicted different score).

Out of all generated texts, all three LLM-based metaevaluators agreed (i.e., a total inter-annotator agreement) on the score in 43\% of cases. They agreed the most on assignment of the score of 2 (5,207 samples) and the least on assignment of the score of 1 (153 samples). The majority score distribution is provided in Table~\ref{tab:score_distribution}. In some cases, the automated parsing of the metaevaluation scores from the LLMs' outputs has been difficult (representing one more challenge of using LLMs for scoring), providing ambiguous values resulting in no meaningful majority score. Fortunately, the number of such cases that were unresolvable even by a manual check was not high, where this occurred only in under 0.1\% of cases.

\begin{table}[!t]
\centering
\resizebox{\linewidth}{!}{
\begin{tabular}{c|c|c}
\hline
\textbf{Score} & \textbf{Absolute distribution} & \textbf{Relative distribution}\\
\hline
3 & 4,135 & 0.1641\\
2 & 11,360 & 0.4508\\
1 & 2,813 & 0.1116\\
0 & 6,874 & 0.2728\\
N/A & 18 & 0.0007\\
\hline
\end{tabular}
}
\caption{Distribution of majority metaevaluation scores.}
\label{tab:score_distribution}
\end{table}

To answer our research question, we are pseudo-randomly splitting this data into label-balanced (based on majority metaevaluation score) subsets for training (1,300 samples per class), validation (500 samples per class), and testing (1,000 samples per class). The training and validation splits are used for language model fine-tuning for multiclass classification task, while the testing split is used for evaluation. Since the testing split is balanced based on labels (the same number of samples in the classes), we are using the standard evaluation metrics of Accuracy and macro average of F1 score (MacroF1).

The results in the following section (especially Table~\ref{tab:basemodels}) show that the metric for evaluation of personalization quality of the multilingual texts can be successfully trained (answering RQ1), with the accuracy well above chance (worst-case accuracy of 0.642 vs random-classifier accuracy of 0.25). The Spearman correlation coefficient between the predicted scores and majority metaevaluation scores on test split is above 0.8 for all tested finetuned models, i.e. representing \textbf{strong correlation}.

Although we cannot directly calculate correlation of PerQ to human judgments due to missing human annotations, we provide an estimation in Appendix~\ref{sec:correlation} using similar data of \citep{zugecova-etal-2025-evaluation}. The results indicate strong correlation to human judgments of the proposed evaluation-metric training approach.

\section{Experimental Results}
\label{sec:results}

The following experiment is focused on comparison of multiple language models when trained for the PerQ metric. The selection of base models for fine-tuning covers multiple sizes and architectures to explore the tradeoff between various aspects (classification performance vs size/inference costs). The results are summarized in Table~\ref{tab:basemodels} and Table~\ref{tab:basemodels_costs}.

\begin{table}[!b]
\centering
\resizebox{\linewidth}{!}{
\begin{tabular}{l|cc}
\hline
\textbf{Model} & \textbf{Accuracy} & \textbf{MacroF1}\\
\hline
Gemma-2-2B & 0.6973 & 0.6955\\
Qwen3-4B & 0.6963 & 0.6954\\
Qwen3-0.6B & 0.6888 & 0.6846\\
DeBERTa-v3-Large (0.4B) & 0.6878 & 0.6926\\
Qwen3-1.7B & 0.6770 & 0.6802\\
mDeBERTa-v3-Base (0.2B) & 0.6598 & 0.6581\\
XLM-RoBERTa-Large (0.4B) & 0.6583 & 0.6572\\
XLM-RoBERTa-Base (0.1B) & 0.6420 & 0.6415\\
\hline
random classifier (4-class) & 0.25 & 0.25\\
\hline
\end{tabular}
}
\caption{Comparison of base models for the PerQ metric prediction performance.}
\label{tab:basemodels}
\end{table}

\begin{table}[!t]
\centering
\resizebox{\linewidth}{!}{
\begin{tabular}{l|ccc}
\hline
\textbf{Model} & \textbf{GPU [GB]} & \textbf{Time [s]}\\
\hline
Gemma-2-2B & 7.71 & 130\\
Qwen3-4B & 10.42 & 208\\
Qwen3-0.6B & 2.97 & 60\\
DeBERTa-v3-Large (0.4B) & 2.48 & 46\\
Qwen3-1.7B & 5.47 & 105\\
mDeBERTa-v3-Base (0.2B) & 2.00 & 19\\
XLM-RoBERTa-Large (0.4B) & 2.26 & 16\\
XLM-RoBERTa-Base (0.1B) & 1.76 & 7\\
\hline
\end{tabular}
}
\caption{Comparison of base models for the PerQ metric inference costs.}
\label{tab:basemodels_costs}
\end{table}

There is only small difference in classification performance between the fine-tuned models (up to 5\%). However, there are high differences in the required GPU memory (for half-precision inference) and eventual inference time. Table~\ref{tab:basemodels_costs} indicates the costs for evaluation by using the test-split data. The costs roughly correspond to the model sizes, especially a higher difference can be seen between the models of >2B parameters and the models of <1B parameters.

When comparing prediction performance and inference costs, the Qwen3-0.6B and DeBERTa-v3-Large models seem to offer the best tradeoff. About 0.5B parameters sized models are usually feasible to be run (inferred) on CPU only, thus increasing the PerQ usability even more.

\section{Case Study}
\label{sec:casestudy}

As a case study to showcase usefulness of the proposed evaluation metric, we select the following research questions:
(\textbf{RQ2}) \textit{``Are there differences in multilingual personalization capabilities (quality of personalization) between small (<=10B) and large (>=20B) LLMs?''}, (\textbf{RQ3}) \textit{``Are there differences between personalization of existing text and personalized text generation?''}, (\textbf{RQ4}) \textit{``Are there differences in personalization quality among languages or target platforms?''}.

To answer these research questions, the researchers would most probably ended up with the usage of LLMs to evaluate the personalization quality (due to subjectivity issues, lack of human annotators across languages, replicability issues, etc.). Therefore, we use the previously-described majority metaevaluation scores to answer such questions. We further analyze the inference costs to obtain such scores and compare them with the inference costs of the PerQ metric to obtain predictions of the personalization quality scores. Afterwards, we compare the observations (regarding the stated research questions) from the two approaches, to signify that the same conclusions can be made at significantly lower costs. For the fair comparison, we use only the test split for the analysis. We provide an ablation study in Appendix, showing approximately the same distribution when analyzing majority metaevaluation from the all data (avoiding potential bias due to pseudo-random test-split selection).

In Table~\ref{tab:metaevaluators_costs}, we provide an estimation of LLM inference costs for the metaevaluation of personalization quality using the three LLMs. The estimation is calculated based on the vLLM-accelerated\footnote{\scriptsize\url{https://docs.vllm.ai/en/latest/}} parallel execution (using 2 GPUs of 64GB) \citep{10.1145/3600006.3613165}. Based on the average metaevaluation speed for a single iteration using the whole dataset, mentioned in Section~\ref{sec:perq}, we have calculated the execution time for the test split of 4,000 samples. The results show that even the fastest LLM metaevaluator (Mistral-Small) is almost $6\times$ slower than the slowest fine-tuned PerQ metric (based on Qwen3-4B), while consuming over $12\times$ more GPU memory. When comparing the combined metaevaluation using all 3 LLMs on 2 GPUs, the time comparison indicates $96\times$ speedup of the PerQ metric (still using $12\times$ less GPU memory). It must be noted that the LLMs often failed in providing the evaluation scores, requiring the repetition of the process and sometimes even manual parsing of the scores from the output, which are not taken into account in this estimation (i.e., actual speedup is even higher). This clearly shows the benefits of the proposed trained metric, accelerating the evaluation of personalization quality from hours to minutes, significantly saving resources (contributing to the energy sustainability of the research).

\begin{table}[!t]
\centering
\resizebox{\linewidth}{!}{
\begin{tabular}{l|ccc}
\hline
\textbf{Model} & \textbf{GPU [GB]} & \textbf{Time [s]}\\
\hline
Mistral-Small-3.1-24B-Instruct-2503 & $2\times64$ & $\sim$1242\\
Aya-Expanse-32B & $2\times64$ & $\sim$9000\\
QwQ-32B & $2\times64$ & $\sim$9840\\
\hline
\end{tabular}
}
\caption{Comparison of LLM-based metaevaluation inference costs. The vLLM acceleration using parallel execution fully utilized 2 GPUs of 64GB, the models themselves would require less GPU resources (however prolonging the execution time).}
\label{tab:metaevaluators_costs}
\end{table}

\begin{figure}[!t]
\centering
\includegraphics[width=\linewidth]{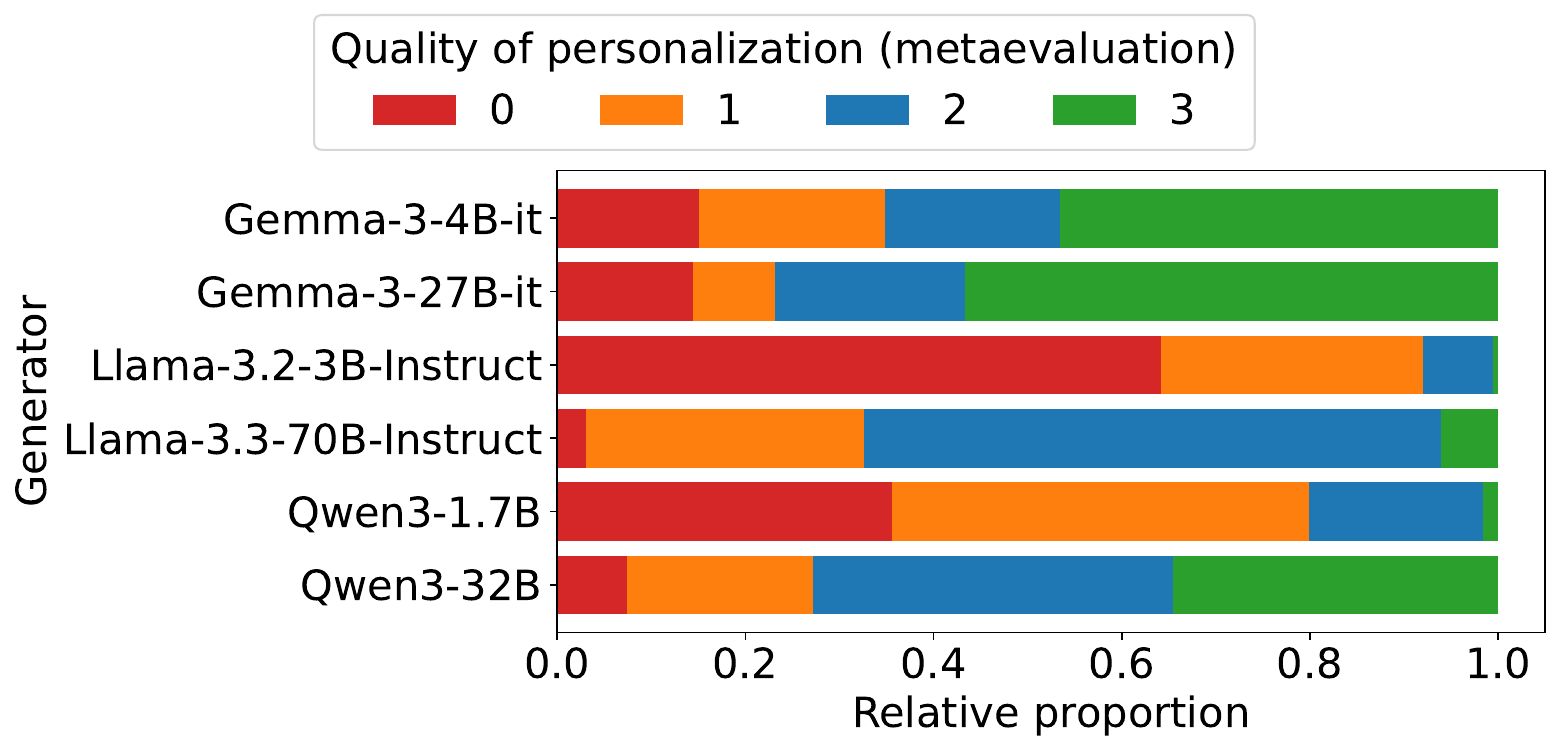}
\includegraphics[width=\linewidth]{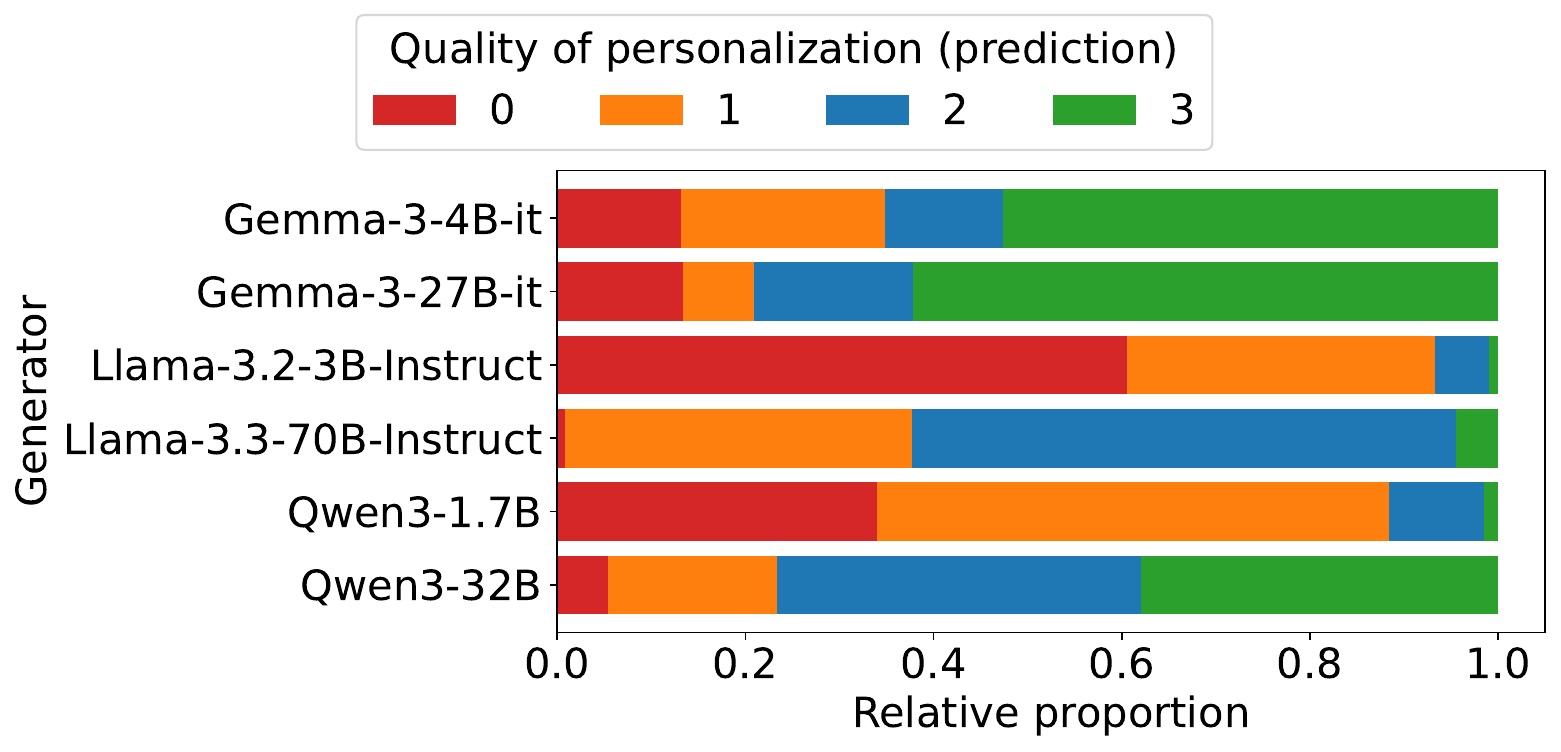}
\caption{Comparison of LLMs personalization capabilities based on the majority metaevaluation scores (top) and the Gemma-based PerQ metric (bottom) for quality of personalization in the test-split texts.}
\label{fig:pergenerator}
\end{figure}

In \figurename~\ref{fig:pergenerator}, we provide per-generator comparison of personalization quality as evaluated by majority metaevaluation (top part of the figure) and by the Gemma-based (as the best performing in Table~\ref{tab:basemodels}) PerQ metric (bottom part of the figure). The distributions of the scores are quite similar using both evaluation approaches. The results (answering RQ2) indicate that smaller variants of the models produce lower quality of personalization than their bigger counterparts. The lowest difference is in the case of Gemma-3 models (almost negligible), and the highest in the case of Llama models (the 3B version generated poorly personalized texts).

\begin{figure}[!t]
\centering
\includegraphics[width=0.85\linewidth]{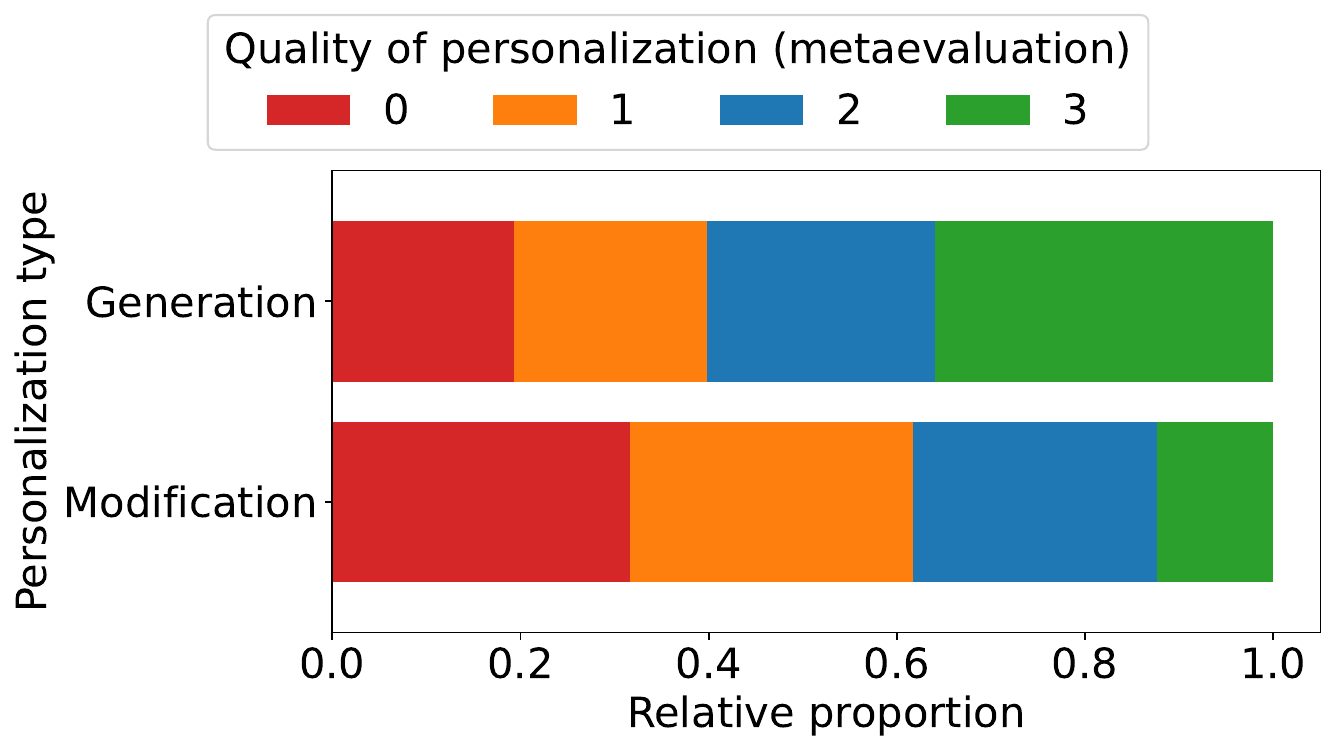}
\includegraphics[width=0.85\linewidth]{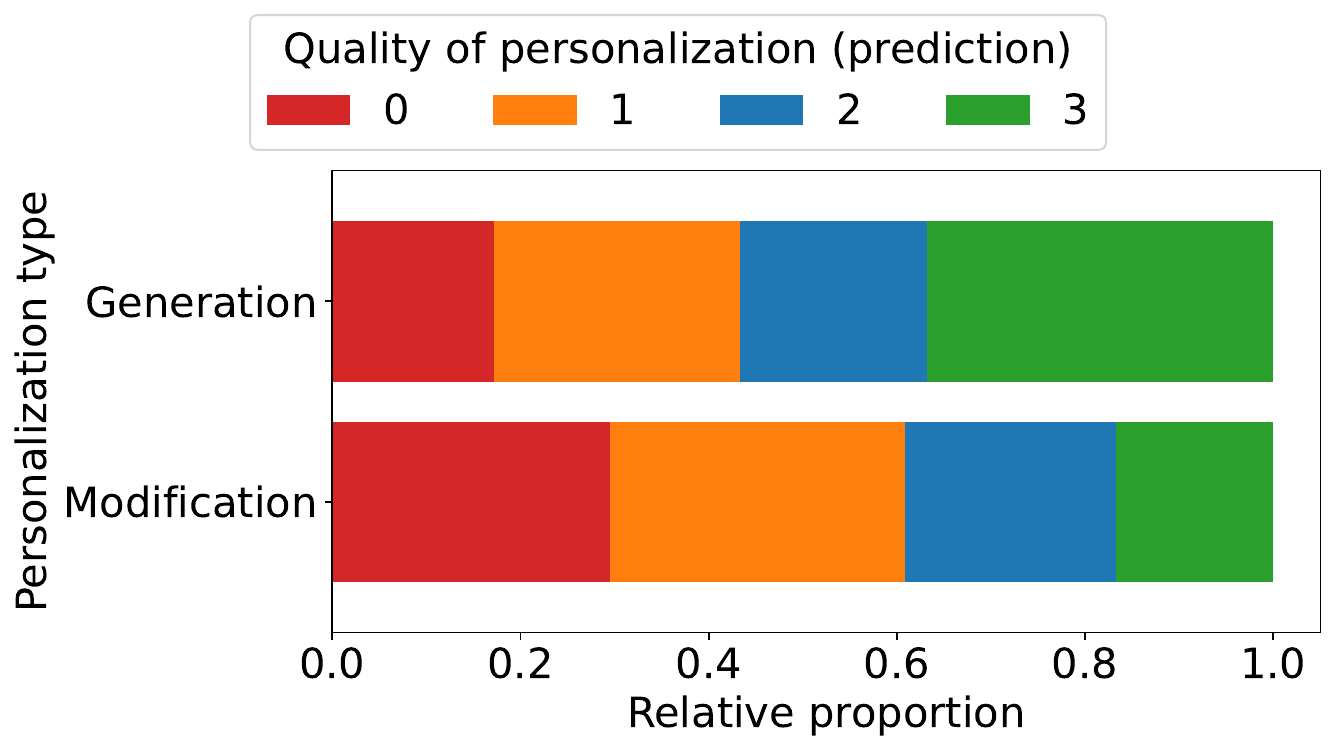}
\caption{Comparison of personalization types based on the majority metaevaluation scores (top) and the Gemma-based PerQ metric (bottom) for quality of personalization in the test-split texts.}
\label{fig:pertype}
\end{figure}

Similarly, to answer RQ3, we provide a comparison between the texts personalized by pure generation and the texts personalized by modification of the existing texts in \figurename~\ref{fig:pertype}. Both approaches of evaluation indicate that the pure generation of the personalized content provides slightly higher quality of personalization (answering RQ3). The results show that in case of generation, the number of texts gaining score of 3 is doubled in comparison to the modification.

\begin{figure}[!t]
\centering
\includegraphics[width=0.9\linewidth]{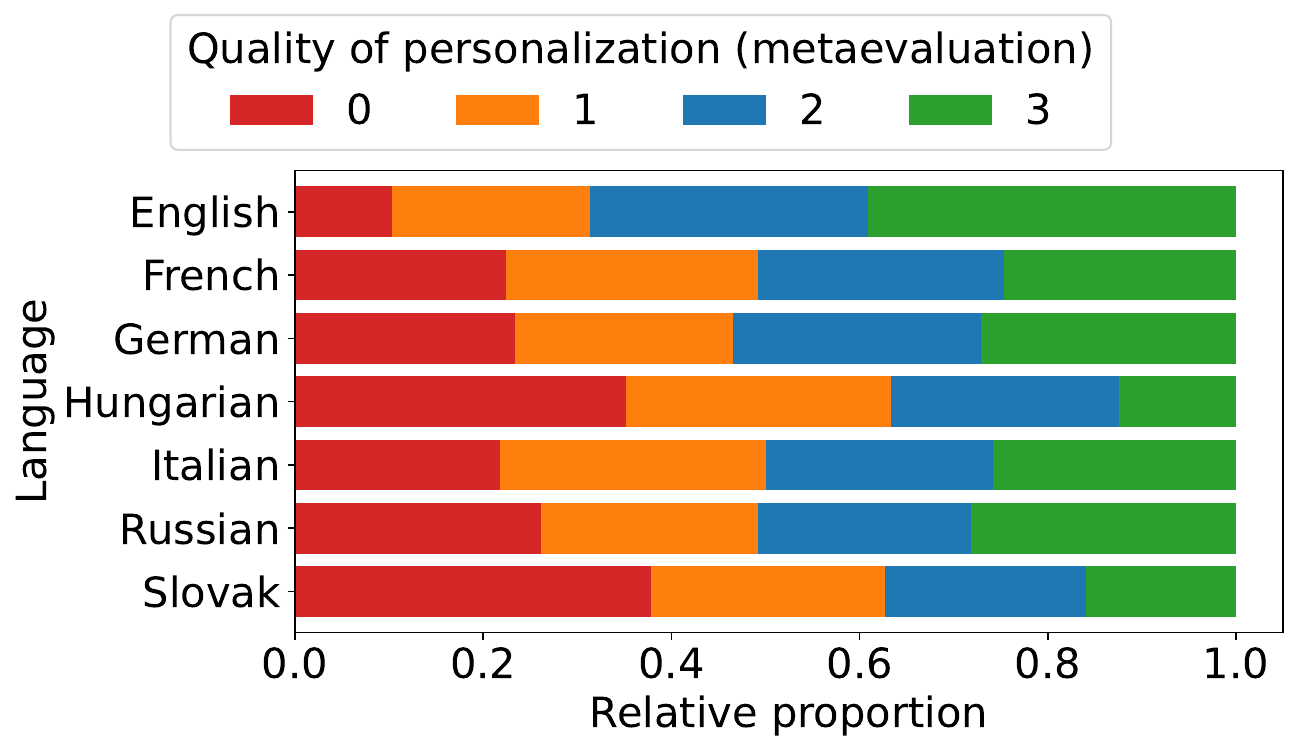}
\includegraphics[width=0.9\linewidth]{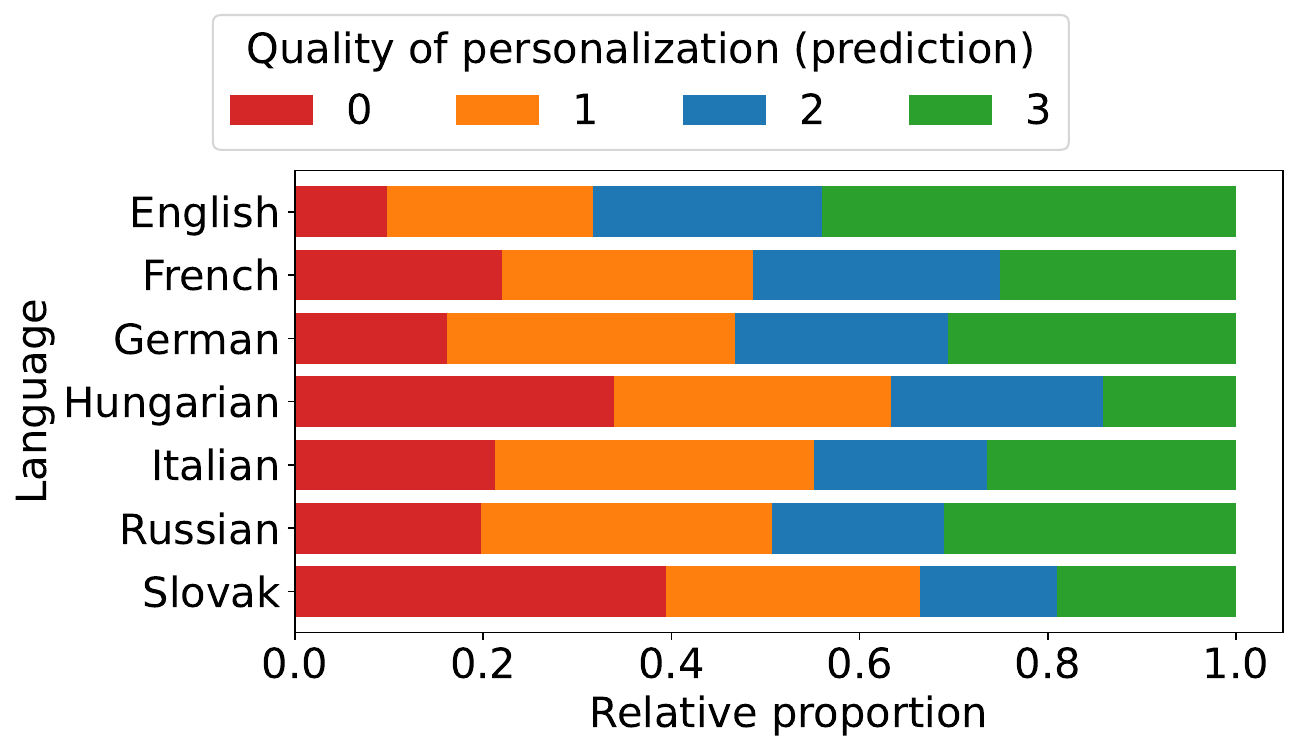}
\caption{Per-language comparison of personalization capabilities based on the majority metaevaluation scores (top) and the Gemma-based PerQ metric (bottom) for quality of personalization in the test-split texts.}
\label{fig:perlanguage}
\end{figure}

\begin{figure}[!t]
\centering
\includegraphics[width=0.9\linewidth]{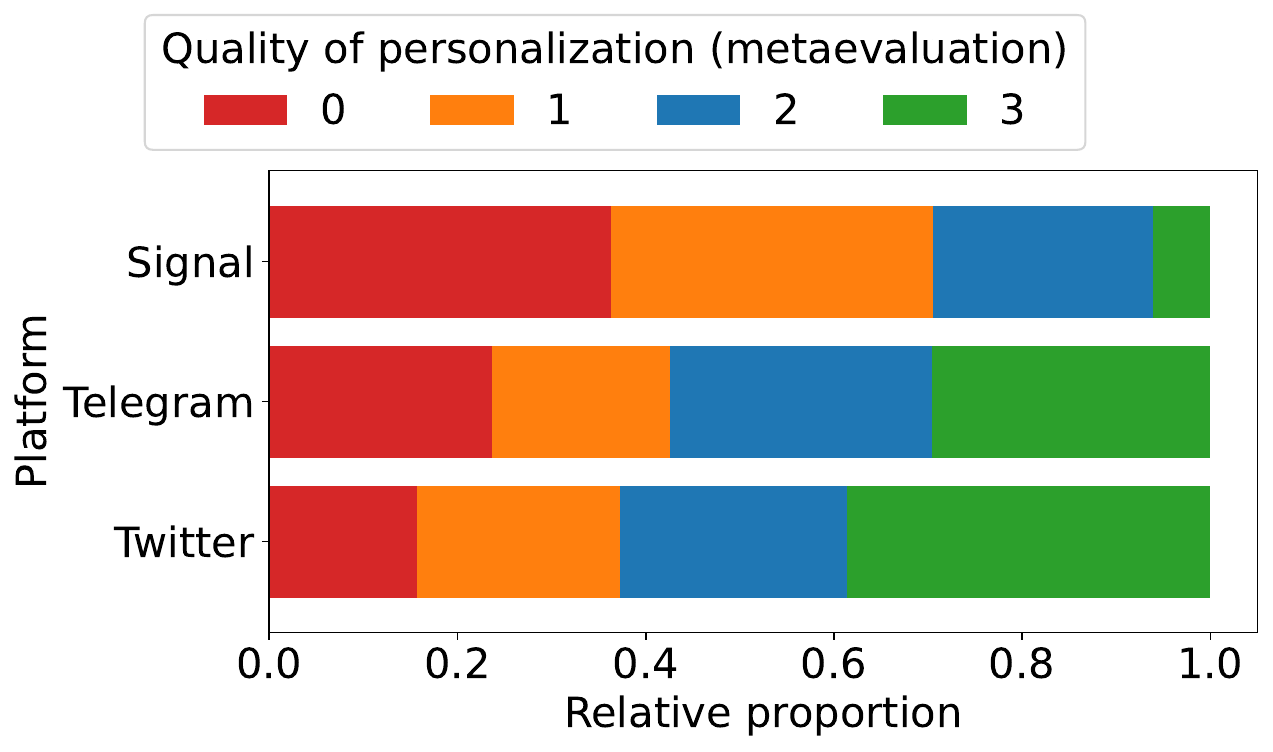}
\includegraphics[width=0.9\linewidth]{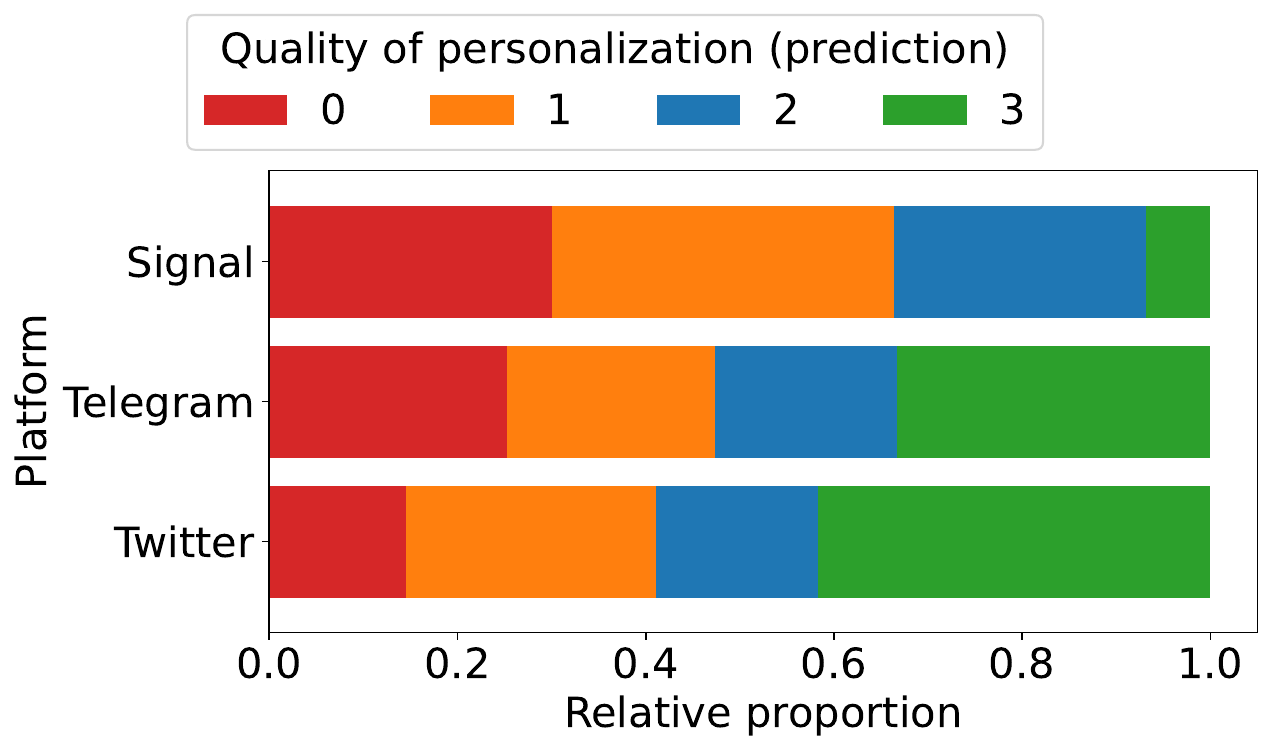}
\caption{Per-platform (i.e., per-target) comparison of personalization capabilities based on the majority metaevaluation scores (top) and the Gemma-based PerQ metric (bottom) for quality of personalization in the test-split texts.}
\label{fig:perplatform}
\end{figure}

When comparing differences among languages (\figurename~\ref{fig:perlanguage}) and among platforms (\figurename~\ref{fig:perplatform}), we can also observe quite similar results from the two evaluation approaches. In both approaches, the Hungarian and Slovak texts are of the lowest personalization quality. The Slovak language contains the biggest proportion of the 0-score texts (the lowest quality) and the Hungarian language contains the lowest proportion of 3-score texts (the highest quality). The English texts are of the best personalization quality (out of the tested languages), indicating the predominant proficiency of LLMs in English, since majority of the pretraining data are English texts.
The results indicate that personalization for Twitter and Telegram platforms was easier for the LLMs than the personalization for the Signal platform (only 6\% of texts scored 3). This effectively answered RQ4.

\section{Conclusions}
\label{sec:conclusion}

The great multilingual capabilities of LLMs offer a way to quickly and easily generate high amount of high-quality texts. However, the measuring the quality of the texts (in various aspects) itself is also often relied on LLMs themselves (due to lack of human evaluators in multiple languages and time-consuming process). However, the usage of high energy-demanding LLMs for evaluation of high amount of texts represents waste of energy and computing resources. In this paper, we have proposed an efficient way of measuring the personalization quality (of targeting to the distribution platform) in a form of the trained metric. The experimental results show a strong correlation of the trained metric prediction to a majority score of 3 LLMs (limiting their internal biases). The speedup can achieve two orders of magnitude, while requiring much less GPU memory.

\section*{Limitations}
\label{sec:limitations}
We have focused on evaluation of personalization quality as defined on our 4-point scale, the evaluation of other aspects of the texts or usage of different evaluation scale might be different.
Our key findings rely on an LLM-based meta-evaluation of personalization quality. Although we have used 3 LLMs to limit the effect of internal biases in evaluation, there can still be some. The study is limited to 7 languages of 4 language-family branches; however, we are unsure about generalization of our findings to other languages. We have limited the generators to 2 variants (smaller and bigger) of 3 SOTA LLMs. The results for other LLMs might differ.

\section*{Ethics Statement}
\label{sec:ethics}
The artifacts and results of this study are intended for research purpose only to evaluate personalization capabilities of existing LLMs. The existing artifacts used in this work have been properly cited and used according their licenses and intended use. We have also checked and followed licensing and terms of use of the used LLMs. AI assistants have not been used for conducting research in any other way than already described in the paper (text generation and meta-evaluation).

\section*{Acknowledgments}
This work was partially supported by the European Union NextGenerationEU through the Recovery and Resilience Plan for Slovakia under the project No. 09I01-03-V04-00068 and partially by \textit{lorAI -- Low Resource Artificial Intelligence}, a project funded by Horizon Europe under \href{https://doi.org/10.3030/101136646}{GA No.101136646}.

\textbf{Computational resources}. We acknowledge EuroHPC Joint Undertaking for awarding us access to Leonardo at CINECA, Italy.

\bibliography{anthology,custom}

\appendix

\section{Computational Resources}
\label{sec:resources}

For the texts generation, we have used 2× A100 64GB GPU, cumulatively consuming approximately 600 GPU-hours. For meta-evaluation, we have used 2x A100 64GB GPU consuming approximately 2000 GPU-hours. For model finetuning, we have used 1× A100 64GB GPU consuming approximately 50 GPU-hours. For other tasks, we have not used GPU acceleration.

\section{Ablation Study}
\label{sec:ablation}

\begin{figure}[!b]
\centering
\includegraphics[width=\linewidth]{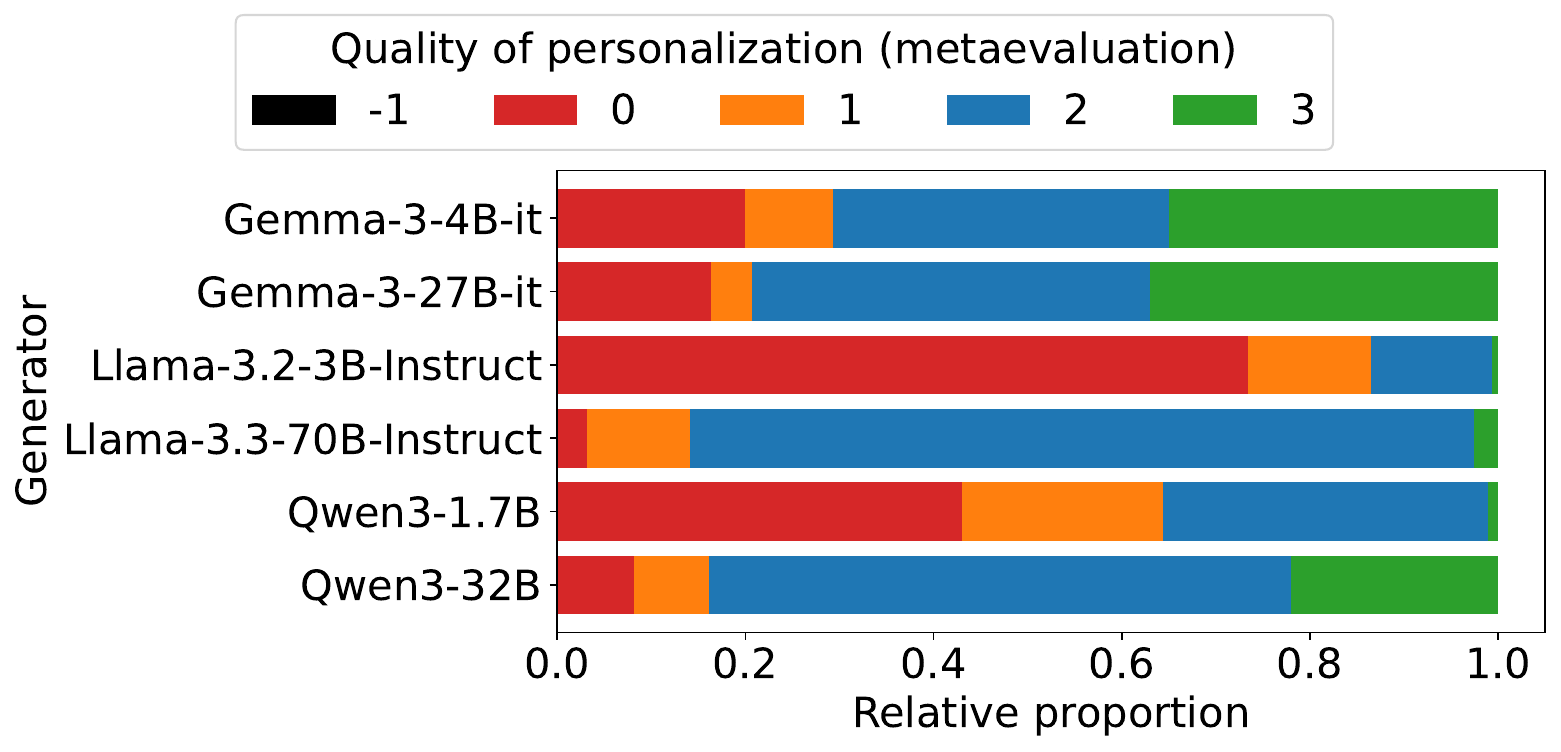}
\caption{Comparison of LLMs personalization capabilities based on the majority metaevaluation scores for quality of personalization in the all texts.}
\label{fig:pergenerator_metaevaluation_all}
\end{figure}

\begin{figure}[!t]
\centering
\includegraphics[width=0.9\linewidth]{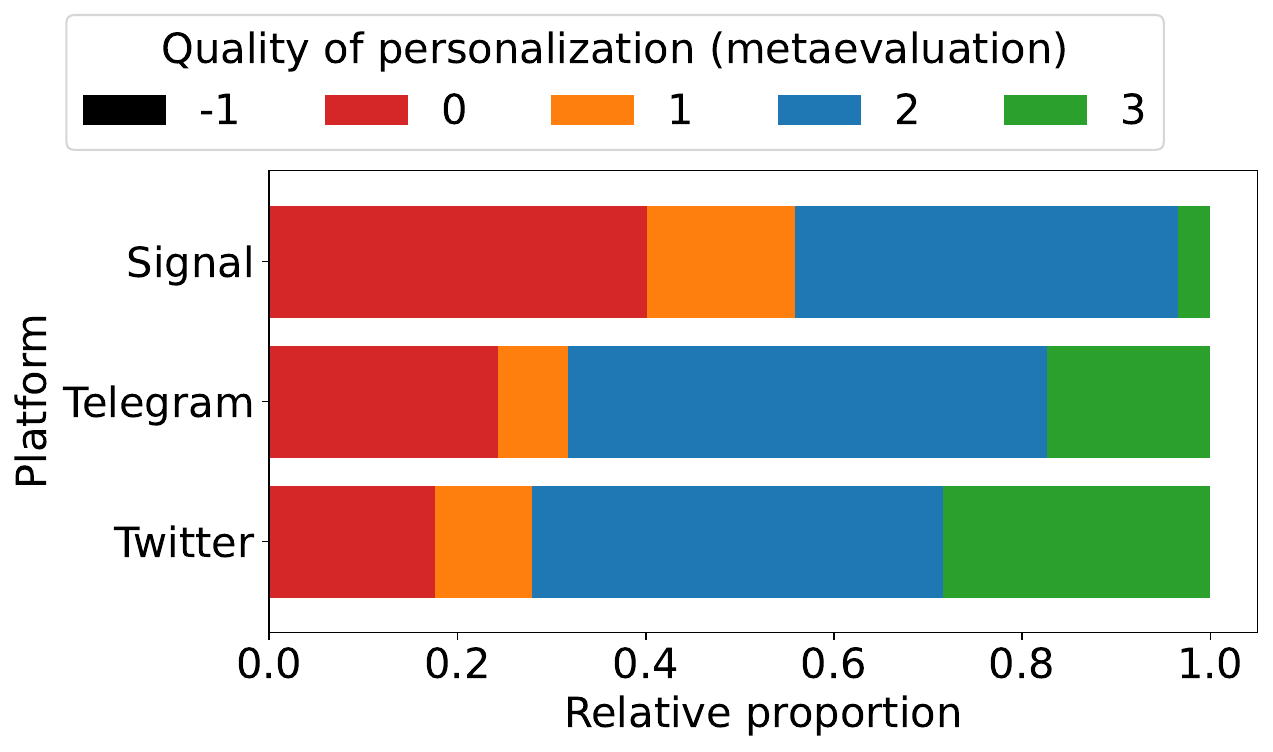}
\caption{Per-platform (i.e., per-target) comparison of personalization capabilities based on the majority metaevaluation scores for quality of personalization in the all texts.}
\label{fig:perplatform_metaevaluation_all}
\end{figure}

Since the analysis of majority-based metaevaluation of personalization quality has been done only on test-split portion of the dataset (to be fairly comparable to the trained PerQ metric), the results might be affected by pseudo-random selection of test split. To analyze whether the results might be different for other selections, we provide the same comparisons of various aspects (as those in the main body of the paper) for majority metaevaluation scores using the whole dataset in
\figurename~\ref{fig:pergenerator_metaevaluation_all} -- \figurename~\ref{fig:perplatform_metaevaluation_all}. This analysis proves that the results are \textbf{not significantly affected} by the selected test-split data.

\begin{figure}[!b]
\centering
\includegraphics[width=0.85\linewidth]{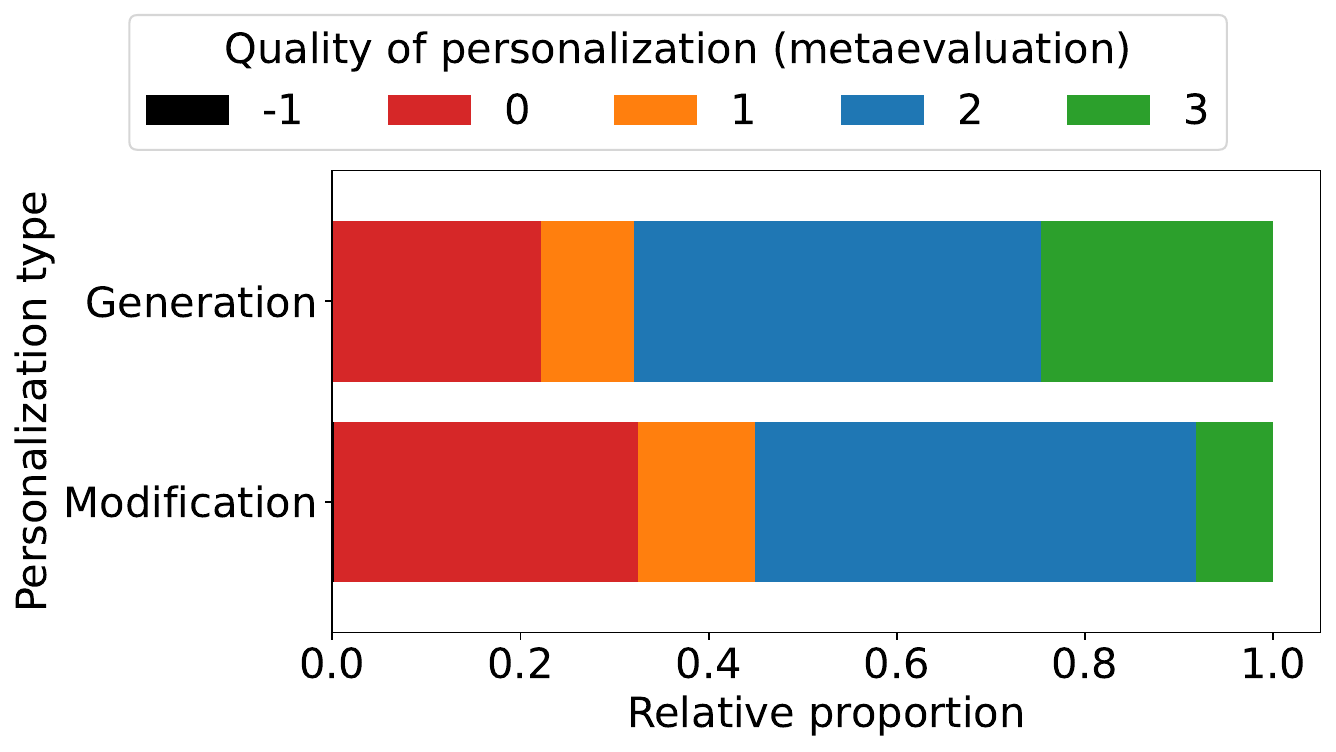}
\caption{Comparison of personalization types based on the majority metaevaluation scores for quality of personalization in the all texts.}
\label{fig:pertype_metaevaluation_all}
\end{figure}

\begin{figure}[!t]
\centering
\includegraphics[width=0.9\linewidth]{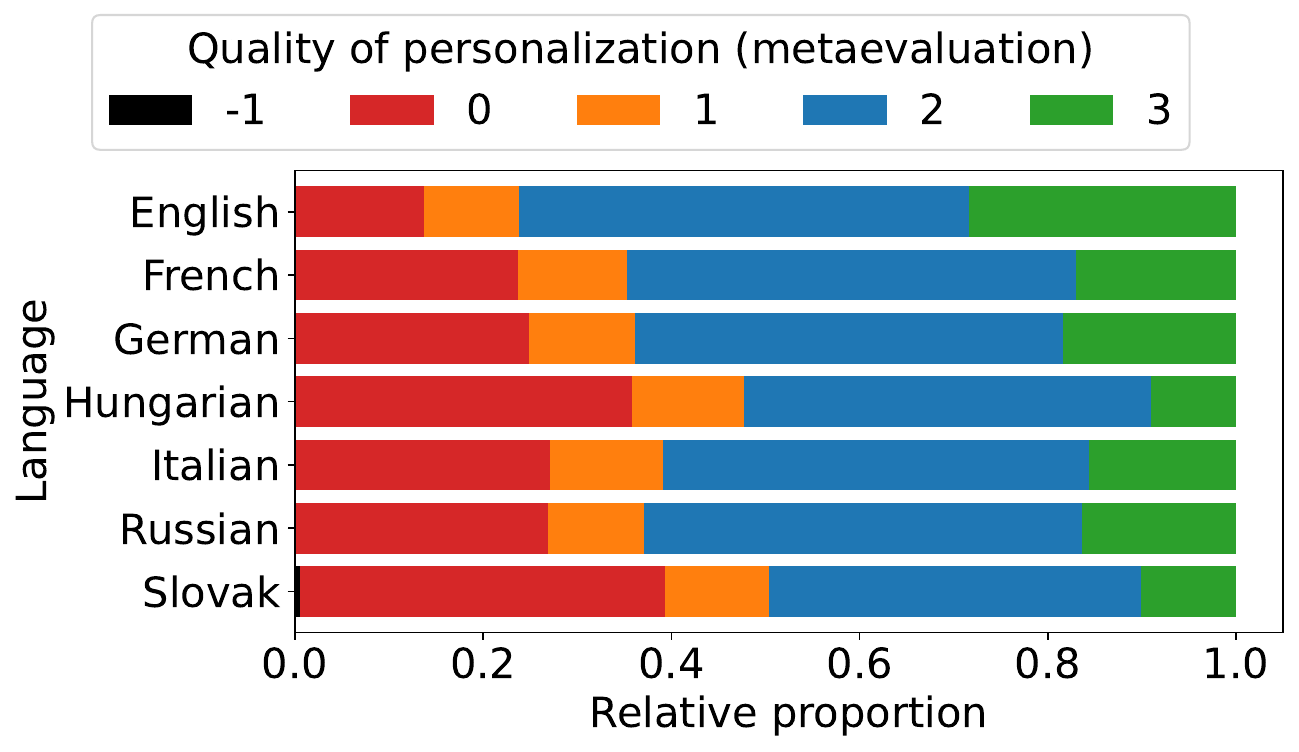}
\caption{Per-language comparison of personalization capabilities based on the majority metaevaluation scores for quality of personalization in the all texts.}
\label{fig:perlanguage_metaevaluation_all}
\end{figure}

\section{Correlation with Human Judgment}
\label{sec:correlation}

Since we have not conducted our own human study due to infeasibility in our research (multilingual settings, time consuming, questionable replicability, costs), we evaluated correlation of the proposed approach with human judgment on similar available dataset (although English only). We have used PerDisNews dataset~\citep{zugecova-etal-2025-evaluation}, which contains LLM-generated personalized English news articles based on various disinformation narratives. Personalization is focused on 7 target groups of people (e.g., students, seniors), which is quite different than our PerQ metric focused on targeting different social-media platforms. However, the evaluation of personalization quality (i.e., the scoring scheme) is similar, also consisting of 4 ordered scores. The dataset includes a subset of 109 samples annotated by 5 human annotators, which is suitable for our evaluation of correlation to human judgments. The dataset also includes metaevaluation of personalization quality by 3 LLMs (although different than we have used), which can be directly used to calculate the majority metaevaluation scores.

Due to differences in personalization itself (target groups vs target platforms), we have trained another personalization-quality evaluation metric using the PerDisNews data that do not have human annotations (2,159 train samples). Although, the majority metaevaluation scores classes were imbalanced, we have used all the data (i.e., without sampling) due to overall low number of samples. It resulted into the accuracy of 0.57 using the evaluation set (human-annotated 109 samples). It completely failed in prediction of score of 2, which was underrepresented in the training set. When compared to majority-based human-annotation scores, the trained metric achieves \textbf{Spearman correlation coefficient of 0.725, indicating strong correlation with human judgment}. We realize this correlation might be different in multilingual settings; however, it still increases confidence in the proposed training approach using majority-based metaevaluation from multiple diverse LLMs.

\end{document}